\documentclass[11pt,a4paper]{article}
\usepackage[hyperref]{acl2021}
\usepackage{times}
\usepackage{latexsym}

\usepackage{multirow}
\usepackage{graphicx}

\usepackage{microtype}

\aclfinalcopy 

\title{UPB at SemEval-2021 Task 8: Extracting Semantic Information on Measurements as Multi-Turn Question Answering}

\author{Andrei-Marius Avram\textsuperscript{1,2}, George-Eduard Zaharia\textsuperscript{1}, \\ \textbf{Dumitru-Clementin Cercel\textsuperscript{1}, Mihai Dascalu\textsuperscript{1}} \\
  University Politehnica of Bucharest, Faculty of Automatic Control and Computers\textsuperscript{1}\\
  Research Institute for Artificial Intelligence, Romanian Academy\textsuperscript{2}\\
  \tt \{andrei\_marius.avram, george.zaharia0806\}@stud.acs.upb.ro\\
  \tt \{dumitru.cercel, mihai.dascalu\}@upb.ro\\
 }

\date{}

\begin{document}
\maketitle
\begin{abstract}
Extracting semantic information on measurements and counts is an important topic in terms of analyzing scientific discourses. The 8th task of SemEval-2021: Counts and Measurements (MeasEval) aimed to boost research in this direction by providing a new dataset on which participants train their models to extract meaningful information on measurements from scientific texts. The competition is composed of five subtasks that build on top of each other: (1) quantity span identification, (2) unit extraction from the identified quantities and their value modifier classification, (3) span identification for measured entities and measured properties, (4) qualifier span identification, and (5) relation extraction between the identified quantities, measured entities, measured properties, and qualifiers. We approached these challenges by first identifying the quantities, extracting their units of measurement, classifying them with corresponding modifiers, and afterwards using them to jointly solve the last three subtasks in a multi-turn question answering manner. Our best performing model obtained an overlapping F1-score of 36.91\% on the test set.
\end{abstract}

\section{Introduction}

Our world revolves around quantities and units of measurement present in all texts, ranging from scientific texts to recipes. Nevertheless, the process of automatically extracting measurements is not trivial, considering that, in most situations, the quantitative structures are ambiguous and are not present in the same area within the text. Therefore, parsing the semantic relations becomes a ubiquitous task, since proper quantity identification leads to transformations towards an easy to follow quantitative summary. Advantages of the previously mentioned process can be found in medical prescriptions \cite{adamo2015automatic}. As such, a system that can robustly and confidently identify medication quantities, measurement units, as well as the medication itself has the potential to become a breakthrough for computer-based medicine and consultations. Another use case resides in ERP systems where proper parsing of resource descriptions facilitates the identification of similar or duplicate items.

The MeasEval - Counts and Measurements competition \cite{MeasEval2021} organized by the 15th International Workshop on Semantic Evaluation (SemEval-2021) creates a new challenge in the area of Natural Language Processing, proposing five subtasks related to span identification, classification, as well as relation extraction, that aim to improve the state of the art for the current field of measurement information extraction. We created a cascaded system to solve the stated problem that is composed of: (1) a subsystem that identifies quantities in the input text; (2) a subsystem that classifies their value modifiers; (3) a subsystem that extracts their measurement unit; and (4) a subsystem that then finds the appropriate measured entities, measured properties, and qualifiers by asking questions related to entity-relations. Three pretrained Transformer-based \cite{vaswani2017attention} language models are experimented for subsystems (1) and (4) of the cascaded system  by fine-tuning them on the specific task: Bidirectional Encoder Representations from Transformers (BERT) \cite{devlin2019bert}, Robustly Optimized  BERT Pretraining Approach (RoBERTa) \cite{liu2019roberta}, and Science BERT (SciBERT) \cite{beltagy2019scibert}.  A character-level bidirectional Long Short-Term Memory (BiLSTM) \cite{hochreiter1997long} architecture was considered for subsystems (2) and (3).

The rest of the paper is structured as follows. The next section presents a series of solutions associated with relation extraction, span identification, and measurement unit identification. The third section outlines our approaches related to the subtasks proposed by the competition. The fourth section presents a performance evaluation of our systems together with an error analysis, while the final section concludes our work and outlines potential future improvements.

\section{Related Work}

\textbf{Span Identification}. \newcite{papay2020dissecting} studied the performance of various models designated for different span identification tasks. Out of them, we mention Conditional Random Fields (CRFs) \cite{lafferty2001conditional}, LSTM cells with CRF, BERT+CRF, LSTM+BERT+CRF, or handcrafted features, usable with any of the previously mentioned models. At the same time, a language model was specifically developed for the span identification tasks, entitled SpanBERT \cite{joshi2020spanbert}, by masking an entire sequence instead of masking a single word in its pretraining process. The authors argued that SpanBERT obtained substantial gains on span selection tasks, such as question answering and coreference resolution.

\textbf{Measurement Unit Identification}. \newcite{Berrahou2013HowTE} proposed a two-step system for search space size reduction, followed by unit extraction from the previously obtained textual fragments. Also, \newcite{hundman2017measurement} presented a hybrid system composed of a CRF that identifies quantities values and their units, followed by a rule-based model to detect their corresponding entities.

\textbf{Relation Extraction.} \newcite{zhang2015relation} adopted a model based on Recurrent Neural Networks (RNN) \cite{cho2014learning} composed of three main elements: an embedding layer, a bidirectional recurrent layer, followed by a max pooling layer that produces the feature vector used for relation classification. RNN-based models were also applied by \newcite{zhangetal2015bidirectional} who adopted BiLSTMs, or by \newcite{xiaoliu2016semantic} who proposed an architecture based on hierarchical RNNs alongside an attention mechanism. Furthermore, several convolutional neural network-based models with various approaches were proposed, for example: multi-level attention \cite{wangetal2016relation}, attention-based context vectors \cite{shenhuang2016attention}, or multi-level features (word, lexical, sentence) \cite{zengetal2014relation}. BiLSTMs are also present in the work of \newcite{lee2019semantic} who implemented a mechanism based on entity-aware attention using latent entity typing. \newcite{jin2020relation} approached the relation extraction task by employing a Graph Neural Network system that modeled each relation as a node and learned the dependencies between the nodes. 

\section{Method}

Our approach on MeasEval consisted of a cascade system composed of individual subsystems for each of the problems in the first two subtasks, and then jointly solving the last three subtasks with a single subsystem.

\subsection{Quantity Identification}

\begin{figure}[hbt!]
    \centering
    \includegraphics[width=0.42\textwidth]{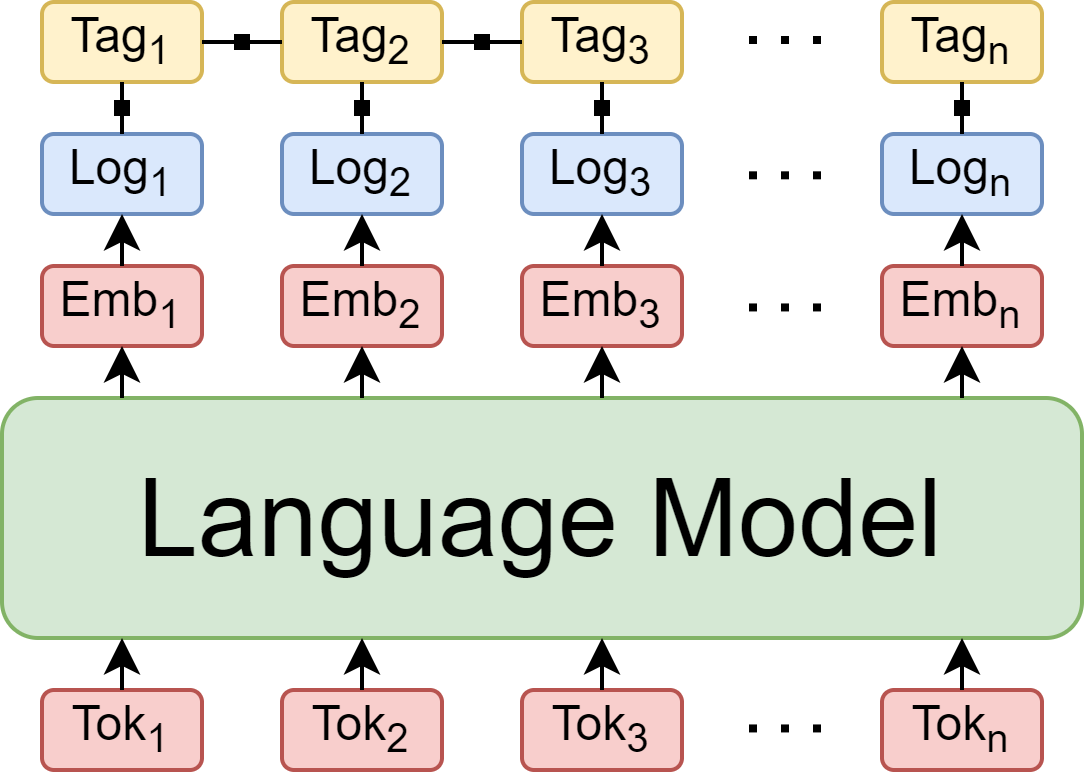}
    \caption{Quantity identification subsystem architecture.}
    \label{fig:quant_identif}
\end{figure}

The subtask of identifying quantities in text was formalized as a sequence labeling problem with Inside–Outside–Beginning (IOB) tags \cite{ramshaw1999text} that were predicted by a pretrained language model with a CRF on top of predicted logits, as proposed by \newcite{avram2020upb}. The architecture is depicted in Figure \ref{fig:quant_identif}.

More formally, we project each output embedding $e_i$ produced by the pretrained language model into probability logits $l_i$ by using a feed-forward network with a ReLU activation as $l_i = ReLU(W^T_le_i + b_l)$, where $W_l$ is the corresponding weight matrix and $b_l$ is the corresponding bias. Then, we model the output conditional probabilities for each tag $y_i$ by using the CRF learning algorithm, as depicted in Eq. \ref{eq:crf_learn}:
\begin{equation}
    p(y|l) = \frac{1}{Z} exp\left\{\sum_{i=1}^{n} {W^T_{y_{i-1}, y_i}l_i + b_{y_{i-1}, y_i}}\right\}
    \label{eq:crf_learn}
\end{equation}
where $W_{y_{i-1}, y_i}$ and $b_{y_{i-1}, y_i}$ are the weight matrix and the bias of the CRF, and $Z$ is a normalization constant such that the probabilities sum up to one.

\begin{figure*}
    \centering
    \includegraphics[width=0.95\textwidth]{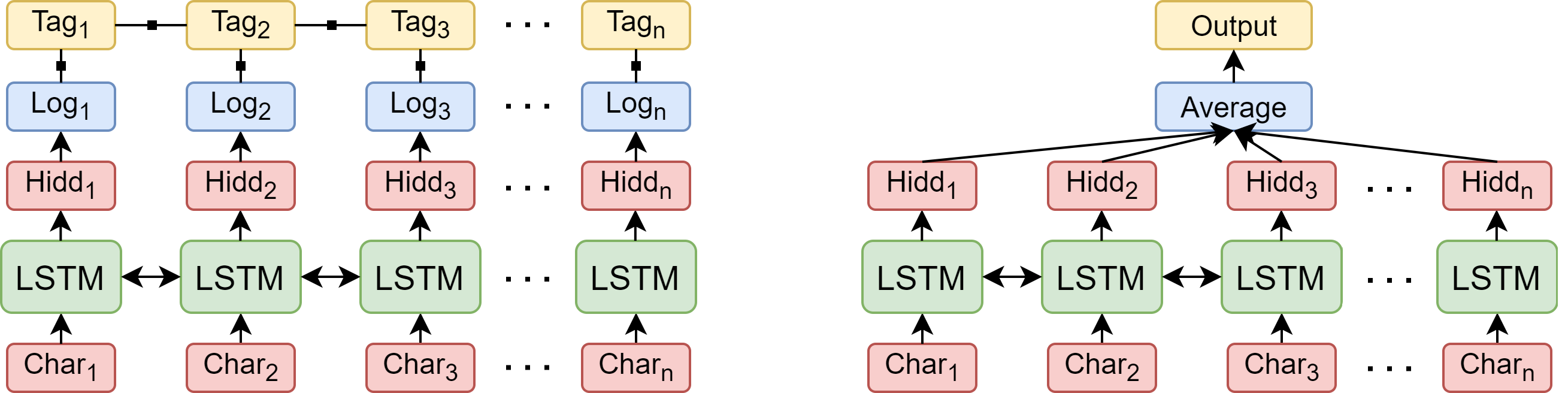}
    \caption{Architectures used in unit extraction (left) and value modifiers classification (right) subsystems.}
    \label{fig:unit_mods}
\end{figure*}

The entire subsystem is trained to maximizing the log-likelihood of the data, while the Viterbi algorithm \cite{forney1973viterbi} is used during inference to find the most likely sequence of tags.

\subsection{Unit Extraction and Value Modifier Classification}

For the second subtask, a character-level BiLSTM extracts the units from quantities and classifies their corresponding value modifiers. We approached the unit extraction in a similar way as the quantity identification, by treating the problem as a sequence tagging; however, the pretrained language model was replaced with BiLSTM cells. Moreover, instead of predicting a label for each character (token), we instead averaged the BiLSTM hidden states and projected their average in an eleven-dimensional vector (i.e., number of possible value modifiers) for the classification. Then, a sigmoid activation function was applied to obtain a vector that contains the probability of the quantity to belong to a class at each index. The architectures used for unit extraction and value modifiers classification are depicted in Figure \ref{fig:unit_mods}.

\subsection{Joint Entity Identification and Relations Extraction}

\textbf{Subtask Grouping.} The last three subtasks were grouped into a single subtask where a pretrained language model was finetuned to jointly identify three elements: the span of the measured entities, the measured properties, and their corresponding qualifiers. The model extracts the relations between the three elements and the previously extracted quantities using a multi-turn question answering (QA) architecture, as proposed by \newcite{li2019entity}. The pretrained language models used for this task were identical to the ones from the quantity identification subtask.

\textbf{Question Templates.} The input to the subsystem is created by appending a question before the text that denotes a possible relation between a given and a target entity. There are a total of six question templates that can be filled with the corresponding entities that cover all the possible relations, as depicted in Table \ref{tab:questions}. Then, the questions are asked in a specific order to correctly identify the relations and the span of the entities. First, starting with a given quantity, the model is asked to identify its measured properties. If a measured property is found, the model marks its span and links it to the quantity with the \texttt{HasQuantity} relation (question 1). Second, the model is asked to identify the measured entity with that corresponding measured property, linking the two with the \texttt{HasProperty} relation (question 2). Third, if no measured property is found for a given quantity, the model is asked to directly identify the measured entity, marking the relation between the measured entity and the quantity directly with \texttt{HasQuantity} (question 3). Finally, once all quantities, measured entities, and  properties are identified, the model is asked to identify corresponding qualifiers and marks the relations accordingly (questions 4-6 in table).

\begin{table*}[hbt!]
    \centering
     \resizebox{\textwidth}{!}{
    \begin{tabular}{cll}
         \hline
          \textbf{\#} & \textbf{Relation Type} & \textbf{Question}\\
          \hline
              1 & HasQuantity & What is the \textit{measured property} of the \textit{quantity} \underline{\hspace{0.8cm}}? \\
              2 & HasProperty & What is the \textit{measured entity} that has the \textit{measured property} \underline{\hspace{0.8cm}} of the \textit{quantity} \underline{\hspace{0.8cm}}? \\
              3 & HasQuantity & What is the \textit{measured entity} that has the \textit{quantity} \underline{\hspace{0.8cm}}? \\
              4 & Qualifies & What is the \textit{qualifier} corresponding to the \textit{quantity} \underline{\hspace{0.8cm}}? \\
              5 & Qualifies & What is the \textit{qualifier} corresponding to the \textit{measured entity} \underline{\hspace{0.8cm}}? \\
              6 & Qualifies & What is the \textit{qualifier} corresponding to the \textit{measured property} \underline{\hspace{0.8cm}}? \\
          \hline
    \end{tabular}
    }
    \caption{Question templates for each relation type.}
    \label{tab:questions}
\end{table*}

\textbf{Model Output.} The architecture proposed in \cite{devlin2019bert} for SQuAD 2.0 \cite{rajpurkar2018know} is employed to create the output of the subtasks; as such, two vectors are used for finetuning: a starting vector $S$ and an ending vector $E$. The probability of token $i$ to be the start of a span is computed as a dot-product between the embedding $T_i$ and the start vector $S$, followed by a softmax applied over all the tokens of the input: $P = softmax(T_i \cdot S)$. An analogous formula computes the end probabilities of a span. Then, we take the indices $i$ and $j$ are taken to compute the most probable span for an entity, where  $i \leq j$ maximizes the sum of log-likelihoods $T_i \cdot S + T_j \cdot E$.  We compare for each query the previously defined maximum sum with the sum of the start and end log-likelihoods of the \texttt{[CLS]} token $s_{null}$ because there can be questions without an answer\footnote{Only measured properties and qualifiers related questions are allowed to not have an answer. Measured entity–related questions must always have an answer.}. If this sum is higher, then there is no such type of relationship for that entity. A threshold added to the $s_{null}$ is considered in order to provide a higher granularity between questions with or without answers, which was tuned on the development set to maximize F1-score.

\begin{figure}[hbt!]
    \centering
    \includegraphics[width=0.48\textwidth]{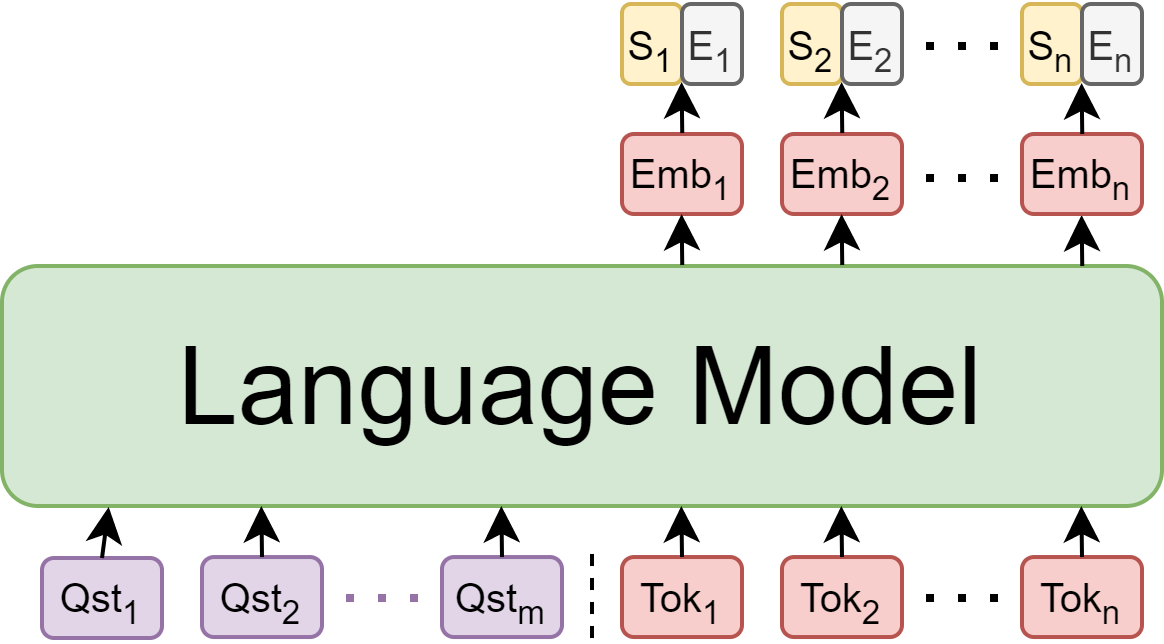}
    \caption{Joint entity recognition and relation extraction architecture as multi-turn question answering.}
    \label{fig:qa_syst}
\end{figure}

Figure \ref{fig:qa_syst} introduces our architecture for entity recognition and relation extraction. The question tokens marked with \texttt{Qst} and the paragraph tokens marked with \texttt{Tok} are fed as input, while the start \texttt{S} and the end \texttt{E} logits are present at output.

\section{Performance Evaluation}
\subsection{Experimental Setup}

\textbf{Dataset Analysis and Processing.} The provided corpus for the competition was quite scarce, counting 298 samples in both train and trial datasets. The corpus contained texts from the scientific domain, counting a total of 7,979 unique words with an average sentence length of approximately 160 words. For training our models, we merged the train and trial subsets and randomly split them into 90\% training and 10\% development.

\textbf{Pretrained Language Models.} An Adam optimizer \cite{kingma2014adam} with a learning rate of 2e-5 was used for training the subsystem of the first and last three subtasks. We experimented with the large versions of BERT and RoBERTa, and with the base version of SciBERT because there are currently no implementations available online of its large variant. Each subsystem was finetuned for 10 epochs; the subsystems that employed large language model variants were trained with a batch size of 2 due to the computational constraints, whereas the subsystems that employed base language model variants used a batch size of 8.

\textbf{BiLSTM Networks.} An Adam optimizer was also employed for training the BiLSTM networks of the second subtask, but with a learning rate of 1e-4. The models were trained for 25 epochs using a batch size of 16. We stacked the LSTM cells two times and used a hidden size of 64, with an embedding of 32 dimensions for the characters. 

\subsection{Results}

The results of each subsystem on the development set are introduced in Table \ref{tab:dev_eval}. SciBERT obtained the highest F1-score on both quantity identification and joint entity and relation extraction, although it is smaller when compared with the other two models. On the second subtasks, the model achieved a reasonable performance, 95.75\% F1-score on unit extraction and 88.94\%  F1-score on value modifier classification.

\begin{table*}
    \centering
    \begin{tabular}{lcccccccccc}
          \multicolumn{1}{c}{} & \multicolumn{2}{c}{\textbf{Avg. Precision}} & \multicolumn{2}{c}{\textbf{Avg. Recall}} & \multicolumn{2}{c}{\textbf{Avg. F1}} & \multicolumn{2}{c}{\textbf{EM}} & \multicolumn{2}{c}{\textbf{Overlap F1}} \\
          \hline
           \textbf{System} & \textbf{Dev} & \textbf{Test} & \textbf{Dev} & \textbf{Test} & \textbf{Dev} & \textbf{Test} & \textbf{Dev} & \textbf{Test} & \textbf{Dev} & \textbf{Test} \\
          \hline
          BERT-related & \textbf{61.10} & 58.69 & 52.28 & 48.05 & 56.63 & 52.84 & 31.61 & 25.58 & 36.72 & 32.69 \\
          RoBERTa-related & 60.04 & \textbf{61.01} & \textbf{56.05} & \textbf{52.66} & \textbf{58.14} & \textbf{56.53} & \textbf{34.98} & \textbf{30.89} & \textbf{39.05} & \textbf{36.91} \\
          SciBERT-related & 56.73 & 54.35 & 54.61 & 46.12 & 55.65 & 49.90 & 30.82 & 23.71 & 35.89 & 30.30 \\
          \hline
    \end{tabular}
    \caption{Results averaged across all five subtasks on the development and test sets.}
    \label{tab:eval_results}
\end{table*}

\begin{table}[hbt!]
    \centering
    \begin{tabular}{lcccc}
         \hline
          \textbf{Subsystem} & \textbf{Precision} & \textbf{Recall} & \textbf{F1} \\
          \hline
            \multicolumn{4}{c}{\textit{Quantity identification}}\\
              RoBERTa-CRF & 90.77 & 92.85 & 91.26 \\
              BERT-CRF & \textbf{91.77} & 93.72 & 92.38 \\
              SciBERT-CRF & 91.60 & \textbf{95.02} & \textbf{93.00} \\ 
          \hline
            \multicolumn{4}{c}{\textit{Unit extraction and value modifier classification}}\\
              Unit Extraction & 96.44 & 95.27 & 95.75 \\
              Value Modifiers & 91.82 & 86.65 & 88.94 \\ 
          \hline
            \multicolumn{4}{c}{\textit{Joint relation extraction and entity identification}}\\
              RoBERTa-QA & 71.04 & 71.26 & 71.14 \\
              BERT-QA & 72.18 & \textbf{71.09} & 71.63 \\
              SciBERT-QA & \textbf{73.81} & 70.71 & \textbf{72.22} \\ 
          \hline
    \end{tabular}
    \caption{Performance analysis on the development set.}
    \label{tab:dev_eval}
\end{table}

The results of the cascaded system are presented in Table \ref{tab:eval_results} that introduces the global precision, recall, and F1-scores averaged across all subtasks, as well as the exact match (EM) and overlap F1 scores\footnote{The overlap F1-score was the metric by which the competition systems were ranked.} between the gold annotations and our predictions. As opposed to the performance of each subsystem on the development set where SciBERT was the best performing model, RoBERTa obtained the highest scores as a whole system, with an overlap F1-score of 39.05\% on the development set and 36.91\% on the test set, outperforming SciBERT with over 6\% and over 3\%, respectively. More surprisingly, SciBERT also obtained a lower score than BERT on both sets, having an overlap F1-score lowered by 2\% and 1\%. We believe that these differences between the scores of RoBERTa and SciBERT were caused by the way the two models were evaluated as stand-alone subsystems or as a whole system.

\subsection{Error Analysis}

\textbf{Quantity Sensitivity.} The main drawback in our approach was that all other subtasks were highly dependent on the quality of the extracted quantities for the first subtask. To exemplify this, let us consider the case where a modifier like \textit{"approximate"} is missed before a quantity; afterwards, it would be impossible to correctly classify its modifiers. Another use case is when the subsystem misses the measuring unit, with the same effect on the unit extractor. Moreover, we noticed that the joint entity and relation extraction was especially sensible to partially identified quantities, producing mostly bad outputs in these cases.

\textbf{Measured Unit Inference.} Another limitation of our approach emerges when the unit extractor does not identify all units in a sequence tagging style. This happened in cases when the unit was split across several places in the quantity, or when it had to be predicted from the context. For instance, the correct unit would be \textit{"m\textsuperscript{2}"} when encountering a  \textit{"300 m x 400 m"} quantity; however, our model found only \textit{"m"} as unit.

\textbf{Long Documents.} Finally, several documents had a longer sequence length than 512 tokens\footnote{Approximately 4\% of the documents had more than 512 tokens for each pretrained language model.} when tokenized, which surpasses the maximum admitted length by the pretrained language models; the workaround was to simply remove the tokens after position 512. However, this solution has the obvious effect of missing identifiable entities that appear after this position. 

\section{Conclusions and Future Work}

This paper introduces our approach that solves all the five subtasks of the 8th task of SemEval-2021 competition in a cascaded manner. First, quantities are identified as a sequence tagging task by using a pretrained language model with a CRF layer. Then,  the measurement units are extracted and the modifiers are classified using BiLSTMs at character level on the identified quantities. Finally, the measured entities, measured properties, and qualifiers are jointly identified, together with their relations, by using a multi-turn question answering approach with hand-crafted questions specific to each relation type. Our best model obtained an F1-score of 36.91\% on the test set. We further emphasized several limitations of our approach and showed that the overall performance was highly sensitive to the quality of the identified quantities.

A possible direction for future work is to test the system using language models that can process longer sequences, such as Longformer \cite{beltagy2020longformer} or BigBird \cite{zaheer2020big}, in order to reduce the effect of missing entities simply due to the sequence length. We also consider creating an ensemble model using several pretrained language models to boost the overall performance, as reported by \newcite{ionescu2020upb}.

\bibliographystyle{acl_natbib}
\bibliography{anthology,acl2021}

\end{document}